\documentclass{isprs} % isprs class modified 23-04-2019 (Dennis Wittich)
\usepackage{amsmath,amssymb,amsfonts}
\usepackage{algorithmic}
\usepackage{geometry} % added 27-02-2014 Markus Englich
\usepackage{graphicx}
\usepackage{subfigure}
\usepackage{setspace}
\usepackage{epstopdf}
\usepackage[labelsep=period]{caption} 
\usepackage[british]{babel} 
\usepackage[hang]{footmisc}
\usepackage{textcomp}
\usepackage{xcolor}
\usepackage[hidelinks, colorlinks=true, linkcolor=blue, citecolor=blue]{hyperref}
\usepackage[authoryear]{natbib}

\begin{document}

% \title{A Probabilistic Drift Correction Approach of 2D SLAM using Geospatial Priors}
\title{A Probabilistic-based Drift Correction Module for Visual Inertial SLAMs}
\date{}

% KAO: Remove extra spacing

% Anonymous submissions, authors' names should not be visible
% \author{
%  Orhan Altan\textsuperscript{1}, Ian Dowman\textsuperscript{2}, Florent Lafarge\textsuperscript{3}, Clément Mallet\textsuperscript{4}, Christian Heipke\textsuperscript{5} }
\author{Pouyan Navard, Alper Yilmaz}

% KAO: Remove extra newline
% Anonymous submissions, authors' affiliations should not be visible
%\address{
%	\textsuperscript{1 }ITU, Civil Engineering Faculty, 80626 Maslak Istanbul, Turkey - (oaltan, tozg, kulur, seker)@itu.edu.tr\\
%	\textsuperscript{2 }Dept.\ of Geomatic Engineering, University College London, Gower Street, London, WC1E 6BT UK - idowman@ge.ucl.ac.uk\\
%	\textsuperscript{3 }Université Côte d’Azur, INRIA – Sophia-Antipolis, France – florent.lafarge@inria.fr\\
%	\textsuperscript{4 }Univ. Gustave Eiffel, IGN-ENSG, LaSTIG – Saint-Mandé, France – clement.mallet@ign.fr\\
%	\textsuperscript{5 }Institute of Photogrammetry and GeoInformation, Leibniz Universit\"at Hannover, Germany - heipke@ipi.uni-hannover.de\\
%}
\address{Photogrammetric Computer Vision Lab, The Ohio State University}

% If the corresponding author is NOT the final author, always add a % space before the subsequent comma, i.e.
% first author name\textsuperscript{a,}\thanks{Corresponding author} , % second author name \textsuperscript{b}, etc.
% thanks to Niclas Borlin 05-05-2016
% information on the corresponding author should not be used any longer and has been commented out
% C. Heipke, Jan 03,2024

% the use of the information of commissions and working groups should not be used any longer and has been commented out
% C. Heipke, Sept. 20,2022
%\commission{XX, }{YY} %This field is optional. If filled, XX and YY should be replaced by adequate numbers. See https://www2.isprs.org/commissions/
%\workinggroup{XX/YY} %This field is optional.
%\icwg{}   %This field is optional.

% KAO: Use times symbol
\abstract{
Positioning is a prominent field of study, notably focusing on Visual Inertial Odometry (VIO) and Simultaneous Localization and Mapping (SLAM) methods. Despite their advancements, these methods often encounter dead-reckoning errors that leads to considerable drift in estimated platform motion especially during long traverses. In such cases, the drift error is not negligible and should be rectified. Our proposed approach minimizes the drift error by correcting the estimated motion generated by any SLAM method at each epoch. Our methodology treats positioning measurements rendered by the SLAM solution as random variables formulated jointly in a multivariate distribution. In this setting, The correction of the drift becomes equivalent to finding the mode of this multivariate distribution which jointly maximizes the likelihood of a set of relevant geo-spatial priors about the platform motion and environment. Our method is integrable into any SLAM/VIO method as an correction module. Our experimental results shows the effectiveness of our approach in minimizing the drift error by $10\times$ in long treverses.
}
\keywords{SLAM, drift, positioning, dead-reckoning}
\maketitle
%\saythanks % added 28-02-2014 Markus Englich

%--------------------------------------------------------
\section{Introduction}\label{introduction}
%--------------------------------------------------------
% % KAO: Sloppy spacing ensures non-overfull lines. Can be removed if this is not an issue.
% \sloppy

Positioning plays a crucial role across a diverse array of applications within the realm of robotics and platform autonomy. At its core, dead reckoning stands out as a leading solution for determining the position of a mobile platform in environments where GPS signals are unavailable or unreliable. This method relies on Markovian state estimation principles, wherein the current position is inferred based on the previous state. However, due to the incremental nature of positioning updates and the inherent noise present in sensor measurements, dead reckoning-based techniques, including both inertial and visual odometry, are susceptible to accumulating errors over time. Consequently, while dead reckoning offers a valuable means of navigation in GPS-denied scenarios, mitigating error accumulation remains a significant challenge in enhancing the accuracy and reliability of such positioning methods.

% Efforts have been made to reduce drift errors using sensor fusion techniques and Bayesian filtering, such as the Kalman filter\cite{bresson2015improving}. For instance, earlier vision-based techniques such as \cite{horn1995continuous,srinivasan1997robot} assist in dead-reckoning errors. Among many others, ORB-SLAM \cite{mur2015orb} has been used frequently and is a monocular SLAM solution that operates in real-time for indoor and outdoor environments. VINS-MONO \cite{qin2018vins} is an alternative SLAM approach that integrates images with inertial measurements to perform relative positioning. The reliability of these and similar methods  decreases when positioning errors inordinately accumulate over time and cause the trajectory to drift from the actual position. Such errors can be diminished with heuristics, such as loop closure \cite{williams2009comparison}. Nonetheless, the problem remains, especially when these heuristics are unmet. This is primarily because the estimation is Markovian, and error accumulates over time \cite{kerl2013dense}, especially in a large-scale environment. 

Various techniques, including sensor fusion and Bayesian filtering such as the Kalman filter \cite{bresson2015improving}, have been employed to mitigate drift errors. Earlier vision-based methods like those proposed by \cite{horn1995continuous,srinivasan1997robot} address dead-reckoning errors, while ORB-SLAM \cite{mur2015orb} offers real-time monocular SLAM functionality for both indoor and outdoor settings. VINS-MONO \cite{qin2018vins}, integrates images with inertial measurements for relative positioning. Despite incorporating heuristics like loop closure \cite{williams2009comparison} to reduce errors such as drift, reliability diminishes when errors accumulate excessively over time, leading to trajectory drift from the true position. This challenge persists, particularly in large-scale environments, due to the accumulative nature of Markovian estimation \cite{kerl2013dense}. There are also some research papaers that study the effect of using GPS in SLAM or VIO setting to improve the positioning accuracy. For example, \cite{kiss2019gps} proposes an augmented version of ORB-SLAM which fuses GPS and inertial data to make the algorithm capable of dealing with low frame rate datasets. \cite{schleicher2009real} studied a real-time hierarchical (topological/metric) SLAM system that is exclusively based on the information provided by both a low-cost, wide-angle stereo camera and a low-cost GPS in large-scale outdoor urban environments. \cite{alsayed20182d} used a Ensemble Multi Layer Perceptron (EMLP) model designed specifically for 2D likelihood SLAM methods. The EMLP model estimates the erors based on the liklihood distribution which makes it independent of the sensor used.

This paper introduces a probabilistic drift correction module tailored for real-time positioning applications. It can easily be plugged in into various relative positioning methodologies. The module operates under the premise that location estimates obtained from the positioning pipeline are inherently uncertain and thus treated as stochastic variables. At each epoch, we define probability density function over the SLAM's estimates to eventually build a unified multivariate distribution.  The module incorporates geospatial priors , as probability distribution, about the characteristic of the platform's motion such as heading, and the surrounding environment into this multivariate framework. Estimating the mode of the multivariate distribution is equivalent to minimizing the accumulated drift error in the SLAM estimate. Section \ref{sec:methodology} discusses our methodology in detail and section \ref{sec:Experiments} evaluates our proposed method in different scenarios. 

%--------------------------------------------------------
\section{Related Work}\label{sec:Related Work}
%--------------------------------------------------------
 
Positioning systems such as VIO\cite{qin2018vins,forster2014svo} and SLAM techniques\cite{mur2015orb,zhou2021lidar,behley2018efficient,keitaanniemi2023drift} usually use camera, Inertial Measurement Unit (IMU)\cite{bai2020high} or other sensor configuration \cite{rivard2008ultrasonic} to estimate the location. For example, VINS-Mono\cite{qin2018vins} is a real-time optimization-based sliding window SLAM framework that leverages from  IMU pre-integration with bias correction and loop detection. Usually, these methods suffer greatly from drift error mainly because their estimation approach is based on Markovian assumption \cite{nobre2017drift,li2020structure,liu2020visual,li2022dr,zhao2020closed}. For instance \cite{botterill2012correcting} overcomes accumulating scale drift problem by recognizing certain objects through the envrionment to and learning their scales to minimize the scale drift issue. There are also positioning systems that uses Global Navigation Satellite System (GNSS), that are very robust to the drift problem, such as GPS which can perform highly accurate positioning stand-alone or in conjunction with SLAM\cite{schleicher2009real,kiss2019gps}. For instance, \cite{hening20173d} introduces a data fusion method employing an Adaptive Extended Kalman filter (AFK) to estimate the velocity and position of a UAV. It combines data from a LIDAR sensor for local position updates via SLAM, corrected by GPS upon the avialability of the GPS signal, and incorporates input from an Inertial Navigation System (INS) into the Extended Kalman filter. While GPS offers precise positioning, its signal can be significantly degraded or be denied in obstructed settings which creates Non-Line-Of-Sight (NLOS) or multipath effect for GPS commonly seen in scenarios such as outdoors environment with thick tree canopies, indoor environment or high altitude\cite{weiss2011monocular,ovstedal2002absolute,chen2018visual}. 

%--------------------------------------------------------
\section{Methodology}\label{sec:methodology}
%--------------------------------------------------------
Assume a platform is moving and, without the loss of generality, has planar motion. Using SLAM, a set of platform positions denoted by a set of points 
$\Gamma = \langle(x_0,y_0),(x_1,y_1),\ldots, (x_n,y_n)\rangle$ is estimated.
The differential geometric representation of the trajectory $\Gamma$ is generated by taking the first-order derivative of consecutive points: 
$$\Delta\Gamma = \langle(dx_1,dy_1),(dx_2,dy_2),\ldots,(dx_{n-1}, dy_{n-1})\rangle.$$ 
This differential representation removes the dependency on the reference frame selection but contains the drift effect. The geometric differential representation introduces the requirement for a known initial position and orientation as initial conditions to set up a differential system of equations introduced later in the discussion. Using two consecutive epochs from $\Delta\Gamma$, we convert the motion vectors into polar coordinates 
\begin{eqnarray}
    \label{eq1_1}
    \phi_t &=& arctan2(dy_t,dx_t),\\
    \label{eq1_2}
    m_t &=& \sqrt{dx_t^2+dy_t^2},
\end{eqnarray} 
and define angular motion at $t$ as $\alpha_t=\phi_t-\phi_{t-1} \in [-\pi,~\pi]$. $m_t \in \mathbb{R}$ in this representation becomes the motion magnitude at $t$. Due to measurement uncertainty, we assume that at $t$, platform angular velocity, $X_1$, and true magnitude, $X_2$ are random variables with Gaussian distributions $X_1 \sim \mathcal{N}(\mu_1 = \alpha_t,~\sigma_{1}^{2})$ and $X_2 \sim \mathcal{N}(\mu_2 = m_t,~\sigma_{2}^{2})$ respectively. These two distributions represent  observables in our module.

\begin{figure}[ht!]
    % \centering
    \centerline{\includegraphics[scale=0.45]{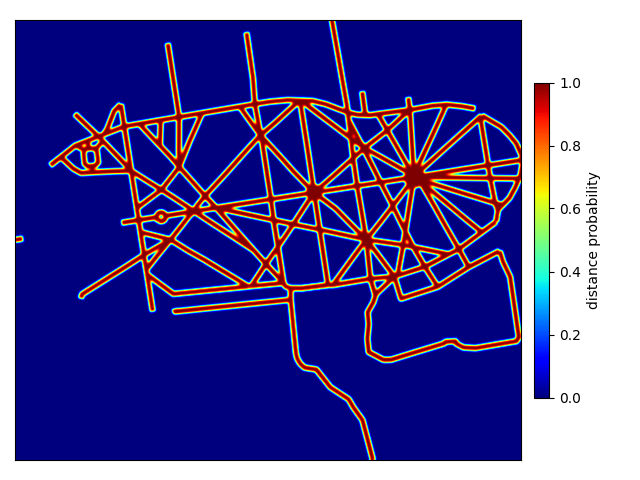}}
    \caption{
        Implicit function representation of vector data through the inverse of the Distance Transform, depicting traversable paths with probabilities favoring center paths and lower values off-center. Maximum allowed distances are truncated akin to road width constraints in GIS-defined traversable paths.
    }
    % \caption{The implicit function representation of the vector data in the form of the inverse of the Distance Transform. The colored regions refer to the traversable paths where higher probabilities are assigned to the center of the paths and lower values to the off-center. Note that the maximum allowed distances are truncated because the platform cannot move beyond the limits of the traversable path defined in GIS. This is akin to the width of a road.}
    \label{fig:1}
\end{figure}

Spatial knowledge of the environment or object motion, represented by additional random variables, can significantly aid in reducing drift error when integrated into our module. Traversable regions within the spatial coverage, such as roads or walkways, are particularly useful for correcting drift errors and is obtained from sources like \cite{OpenStreetMap}. These regions are typically represented as vectors in Geospatial Information Systems (GIS), which is rasterized for analysis. However, due to the complexity of the vector shapes, using an explicit function for parametric distribution isn't always feasible. To address this, in our paper, we convert rasterized paths into an implicit function, akin to those used in fluid dynamics \cite{sussman1999efficient} for analyzing moving fronts. This implicit function, depicted in Figure \ref{fig:1}, is based on the distance function ($\mathcal{D}$), yielding a distribution where $X_3 \sim \mathcal{N}(\mu_3 = 0,~\sigma_{3}^{2})$ where $\sigma_{3}$ modeling positioning uncertainty in the GIS map.

% Spatial knowledge  about the environment or the object motion in the form of additional random variables can help further reduce drift error by including them in our module. For instance, traversable regions in the spatial coverage, where the platform moves, have a great utility in correcting drift errors. In Geospatial Information Systems (GIS), traversable regions, such as roads, walkways, etc., are represented as vectors that can be obtained from different sources, such as OpenStreeMap (OSM). These vectors trace a path on a map and can be rasterized. Considering the complex shapes the vectors can trace, having an explicit function that can be used in a parametric distribution is not always applicable. In our paper, to facilitate generating a random variable for traversability in spatial coordinates, we convert the rasterized paths into an implicit function that is frequently used to analyze moving fronts in fluid dynamics \cite{sussman1999efficient}. We use the distance function ($\mathcal{D}$) with the following property $\mathcal{D}(x,y)\mapsto\mathbb{R}$ that is depicted in Fig. \ref{fig:1} as the implicit function. This formulation results in the following distribution $X_3 \sim \mathcal{N}(\mu_3 = 0,~\sigma_{3}^{2})$ where $\sigma_{3}$ models the positioning uncertainty in the GIS map. 

Another example of a useful assumption is motion constraints, such as the heading preservation, $H$, that is added to the random variable set. This constraint assures that the current heading direction does not change radically between consecutive epochs. The random variable associated with this constraint is defined to follow normal distribution like the previous motion-related random variables $X_4 \sim \mathcal{N}(\mu_4 = 0,~\sigma_{4}^{2})$, and it shares the same calculation procedure as in \eqref{eq1_1}. 

\begin{figure}[t!]
    \centering
    \includegraphics[scale=0.12]{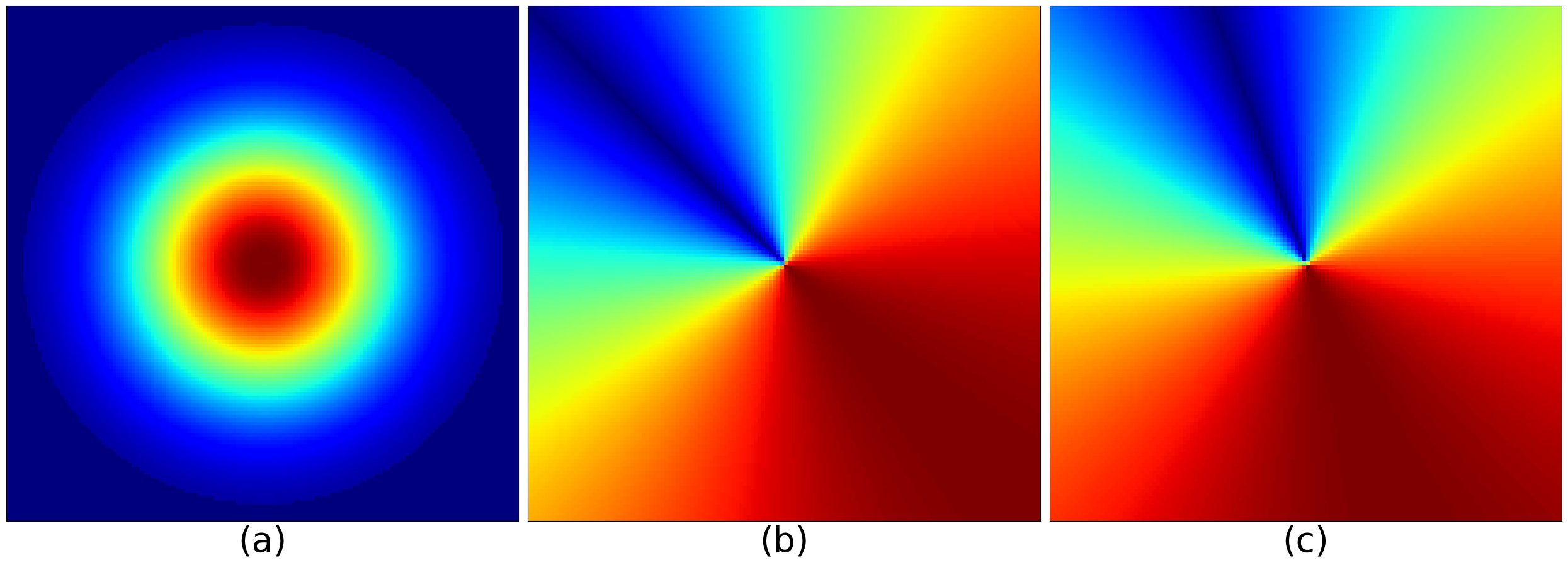}
    \caption{Spatial probabilities at $\Gamma(t)$ where red and blue denote high and low probability values, respectively. Random variables (a) motion magnitude $X_2$, (b) motion angle $X_1$ (c) heading preservation $X_4$. }
    \label{fig:2}
\end{figure}

In our multivariate recipe, each distribution discussed above contains useful beliefs about the environment and the traversed trajectory at time $t$, providing a practical utility to exploit these priors simultaneously. Considering independence between random variables, the multivariate distribution can be written as:
\begin{equation}
    \label{eq2}
    P = \prod_{i=1}^4 P(X_i~|~\mu_i,~\sigma_i)
\end{equation}
where each term in this equation generates a spatial distribution that is visualized in Fig. \ref{fig:2}.
Having formed the joint prior distribution, recursive filtering is  applied to yield a posterior probability density function. This procedure is statistically equivalent to  estimating the position as the distribution mode. Considering the model in \ref{eq2}, the maximization is more tractable when a negative log-likelihood approach converts the problem to a minimization problem. Therefore, \eqref{eq2}  becomes: 
\begin{equation}
    \label{eq3}
    P = -\sum_{n=1}^{4} w_i \cdot\ln{P(X_i~|~\mu_i,~\sigma_i)}, 
\end{equation}
where $w_i \in (0,~1]$ is used to weigh the contribution of each constraint depending on their reliability in obtaining the corrected position estimate. The solution to this equation is computed by gradient descent.

%--------------------------------------------------------
\section{Experiments}\label{sec:Experiments}
%--------------------------------------------------------
The experiments in this section use VINS-MONO, a visual and inertial SLAM for the relative positioning pipeline, to generate $\Gamma$. The data is recorded on a smartphone using the MarsLogger tool \cite{huai2019mobile} that records images at 30Hz and IMU measurements at 100Hz synced to the same clock source. We collected three scenarios containing trajectories with sharp turns, long straight traverses, and loop closures to test the proposed drift correction module to SLAM. 

To resolve for the mode of the proposed multivariate distribution depicted in \ref{eq3}, we used PyTorch \cite{paszke2017automatic} to perform Stochastic Gradient Decent \cite{bottou1991stochastic} optimization with learning rate of 0.01. For taking the derivative of the distance transform grid depicted in \ref{fig:1}, bilinear interpolation is used to estimate the derivatives for non-discrete coordinates where derivative is not directly available from the grid.

\begin{figure*}[ht!]
    \begin{tabular}{c}
    \centerline{\includegraphics[scale=0.27]{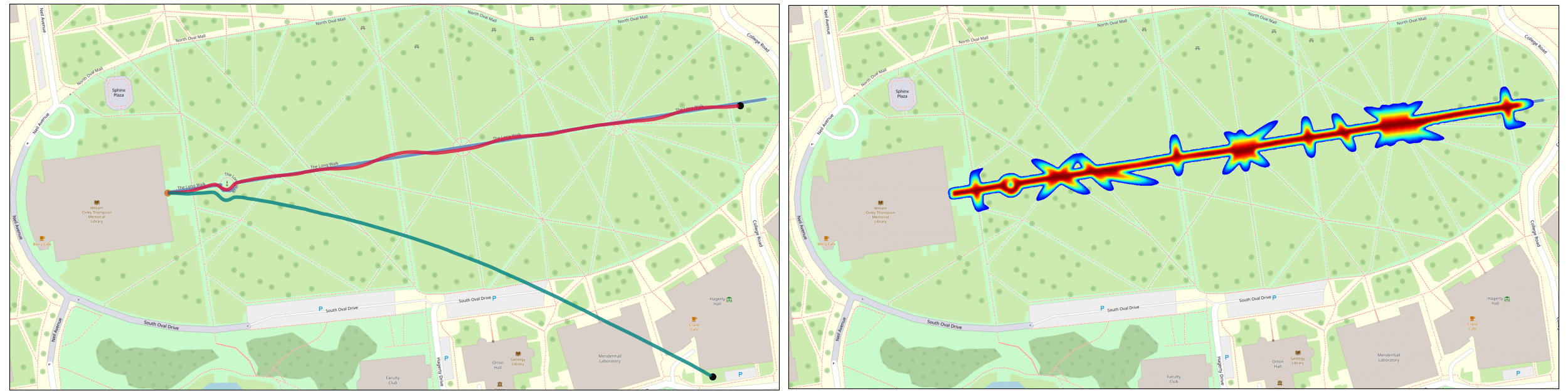}}\\{\scriptsize(a)}\\
    \centerline{\includegraphics[scale=0.27]{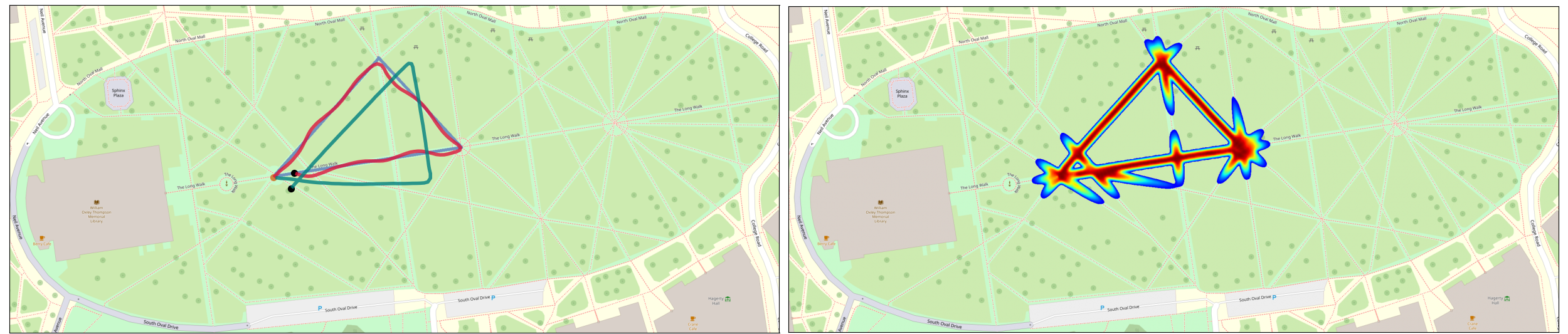}}\\{\scriptsize(b)}\\
    \centerline{\includegraphics[scale=0.27]{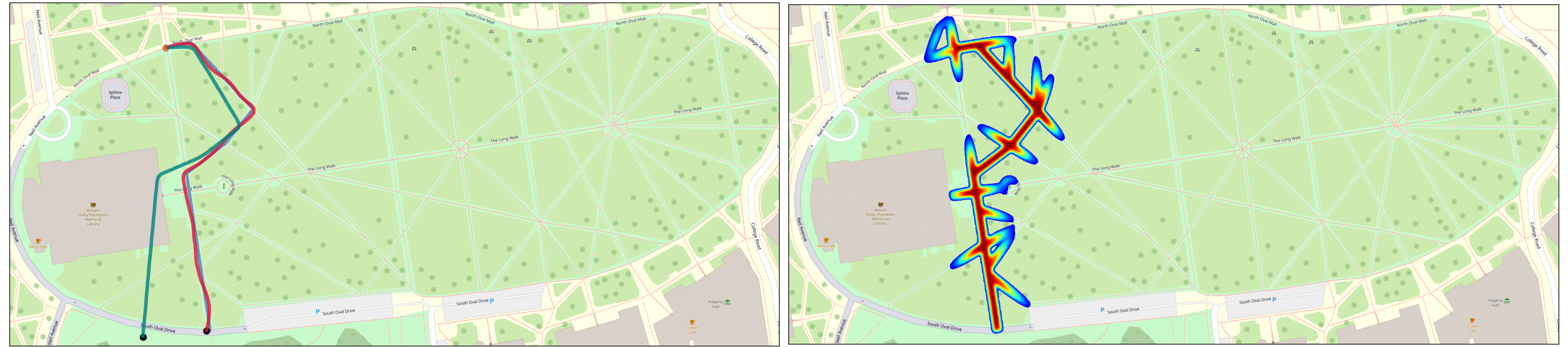}}\\{\scriptsize(c)}
    \end{tabular}
    \caption{
        Comparison of reference, proposed, and SLAM trajectories (blue, red, and green) with start and end points highlighted (orange and black), alongside spatial distributions generated by the proposed method on the right had side.
    }
    \label{fig:3}
\end{figure*}

The proposed trajectories in Fig. \ref{fig:3} qualitatively follow the ground truth in all testing scenarios. However, the instability in the proposed trajectories occurs due to the competing effect of having different terms in the mixture model. One can perform an ablation study to choose the correct distribution weights $w_i$ and the model's uncertainty parameters $\sigma_i$.

Table \ref{tabel1} compares the closing distances between the estimated trajectories, using our module and SLAM only, and the reference trajectory generated from the GPS readings. The labels in the first column of the table respectively correspond to the labels in Fig. \ref{fig:3}. We can see that the drift correction module successfully overcomes the drift error by ten to twenty times for long traverses. For the loop closure example, the results are similar, yet as observed from the qualitative results, the maximum error is higher.

\begin{table}[h]
\centering
\caption{
    Computed trajectory's closing distance to reference trajectory, reported in meters up to decimeter error, accounting for GPS sensor reliability based on specifications.
}
% \caption{Closing distance between the computed and reference trajectory reported in meters. Results are reported up to decimeter error due to the reliability of the GPS sensor based on specs.}
\begin{tabular}[t]{crr}
\hline
Fig \ref{fig:3} part & Our Method (m.)& SLAM only (m.)\\
\hline
(a) & 15.8  & 171.8\\
(b) & 12.7  & 12.6\\
(c) & 2.3 & 40.9\\
\hline
\label{tabel1}
\end{tabular}
\end{table}%

%--------------------------------------------------------
\section{Conclusion}
%--------------------------------------------------------

This paper proposes a differential geometric correction module for the drift error in localization and navigation pipelines, such as SLAM or VIO. The drift error results in significant deviation, especially when the platform traverses large-scale environments. The proposed drift correction module formulates the platform positioning problem as a mode seeking of a multivariate probability distribution generated from differential geometric observations. This strategy and the independence of the module from the SLAM approach make the proposed drift correction to be used by most SLAM algorithms without difficulty. The results for various scenarios show the module successfully corrects the drift.

% This paper introduces a novel solution aimed at rectifying the prevalent issue of drift error within localization and navigation pipelines, including SLAM (Simultaneous Localization and Mapping) or VIO (Visual-Inertial Odometry). Drift error poses a significant challenge, particularly in large-scale environments, where it can lead to substantial deviations from the true path. The proposed drift correction module addresses this by employing a differential geometric framework. It conceptualizes the platform's positioning as a mode-seeking problem within a multivariate probability distribution derived from the differential observations. One of the key advantages of this approach is its independence from any specific SLAM methodology, enabling seamless integration with various SLAM algorithms. Results obtained across different scenarios demonstrate the module's effectiveness in successfully mitigating drift and improving overall localization accuracy.

%--------------------------------------------------------
\clearpage
	\begin{spacing}{1.17}
		\normalsize
		\bibliography{references} 
	\end{spacing}
\end{document}